\PassOptionsToPackage{table}{xcolor}
\PassOptionsToPackage{authoryear,sort&compress,round}{natbib}
\documentclass[11pt,a4paper,twocolumn,logo]{googlecloud}
\usepackage{natbib}
\usepackage{multirow,arydshln}
\usepackage{listings}
\usepackage{algorithm,algpseudocode}
\usepackage{xspace}
\definecolor{cathdr}{HTML}{F0F3F7}
\definecolor{ooblue}{HTML}{3367B2}
\newcommand{\oohfull}{Object-Oriented Harness\xspace}

\newcommand{\ooh}{\mbox{\textsc{OO-Harness}}\xspace}
\newcommand{\mh}{\textsc{Meta-Harness}\xspace}
\newcommand{\ahe}{\textsc{AHE}\xspace}
\newcommand{\aah}{\textsc{AAH}\xspace}
\newcommand{\seam}[1]{\texttt{#1}}
\hypersetup{
  pdftitle={Object-Oriented Harness: Evolving Agent Harnesses through Object-Oriented Design},
  pdfauthor={Gaotang Li, Bhavana Dalvi Mishra, Yifei Ming, Yanfei Chen, Zhongying CuiZhu, Lesly Miculicich, Long T. Le, Rui Meng, Chun-Liang Li, Hanghang Tong, Chen-Yu Lee, Tomas Pfister}}

\title{Object-Oriented Harness: Evolving Agent Harnesses through Object-Oriented Design}
\author[1,*]{Gaotang~Li}
\author[2]{Bhavana~Dalvi~Mishra}
\author[2]{Yifei~Ming}
\author[2]{Yanfei~Chen}
\author[2]{Zhongying~CuiZhu}
\author[2]{Lesly~Miculicich}
\author[2]{Long~T.~Le}
\author[2]{Rui~Meng}
\author[2]{Chun-Liang~Li}
\author[1]{Hanghang~Tong}
\author[2]{Chen-Yu~Lee}
\author[2]{Tomas~Pfister}

\affil[1]{University of Illinois Urbana-Champaign}
\affil[2]{Google Cloud AI Research}

\begin{document}
\raggedbottom
\begin{abstract}
Agent harnesses are executable code that specifies the prompts, tools, and control flow around a language model.
A single global harness can struggle with heterogeneous tasks that require different execution strategies.
In this work, we introduce \oohfull (\ooh), a harness optimization framework that evolves a hierarchy of executable harness classes using object-oriented design.
During training, an evolution agent uses execution records to create a family of specialized harnesses, each implemented as a subclass of the base or an existing specialized harness via inheritance.
At test time, a harness router uses routing lessons accumulated during training to select a specific harness for each task.
Across three long-horizon benchmarks spanning business workflows, workspace tasks, and tool use, \ooh achieves the best performance among the compared methods, with an absolute improvement of up to $15.79\%$ over the base harness.
Extensive ablation studies examine the contributions of its core components.
\end{abstract}
\maketitle
\begingroup
\renewcommand{\thefootnote}{\fnsymbol{footnote}}
\footnotetext[1]{This work was done during Gaotang Li's internship at Google Cloud AI Research.}
\endgroup

\begin{figure}[!t]
  \centering
  \includegraphics[width=\linewidth]{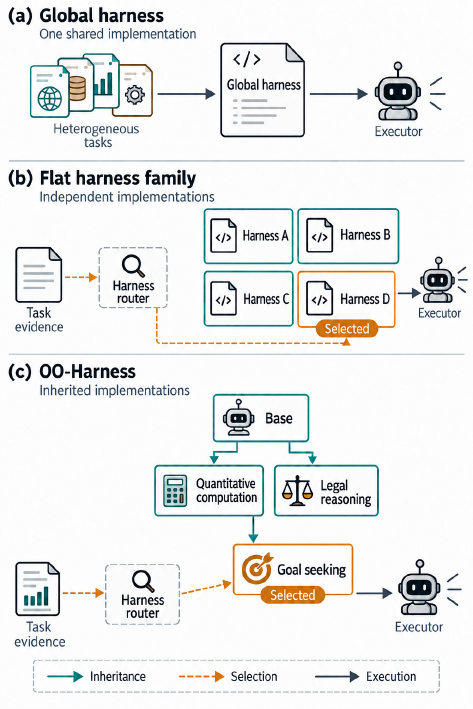}
  \caption{\textbf{Three ways to organize harness adaptation.}
  (a) A single global harness serves all tasks.
  (b) A router selects among independent harness implementations.
  (c) \ooh selects among specialized harnesses that share code through
  inheritance.
  }
  \label{fig:motivation}
\end{figure}

\section{Introduction}
\label{sec:intro}
Generalizing to tasks beyond the training distribution is a central goal
of language-agent research~\citep{zeng2023agenttuning,xi2024agentgym}.
Recent benchmarks illustrate the breadth of tasks agents must handle:
FutureX evaluates agents on forecasting real-world events using external
information~\citep{zeng2025futurex}, while APEX-Agents evaluates
cross-application workflows in investment banking, management consulting,
and corporate law~\citep{vidgen2026apexagents}.
These workloads differ in the tool use, context management, and completion
checks they require. An agent's \emph{harness} implements these choices as
executable code around a language model~\citep{yang2024sweagent}.

Recent approaches automate agent design and harness optimization by
refining executable code and other harness components using execution
feedback~\citep{hu2024adas,zhang2025dgm,lou2026autoharness,lee2026metaharness,lin2026ahe,karten2026continual,nie2026evolvenet,jiang2026harnessevolve}.
Meta-Harness, for example, evolves candidate programs using discovery-set
feedback and selects a global harness based on performance on that
set~\citep{lee2026metaharness}.
Such selection does not guarantee that the same changes remain beneficial
under distribution shift.
Even within a benchmark, Agentic Harness Engineering's component ablations
show that adding evolved memory improves performance on hard tasks but
reduces it on easy ones~\citep{lin2026ahe}.
EvoHarnessBench further shows that adding tools, skills, and specialist
agents can degrade previously effective behavior~\citep{ke2026evoharnessbench}.
These findings motivate specialized harnesses that share common code while
preserving different execution strategies, with a router selecting a harness
for each task.

Because harnesses are executable code~\citep{ning2026codeharness},
object-oriented design offers a natural way to express shared and
specialized behavior. We introduce \oohfull (\ooh), a harness optimization
framework that evolves a hierarchy of executable classes
(Figure~\ref{fig:motivation}). Each specialized harness inherits from the
base or an existing specialized harness, overriding inherited methods or
adding new ones.

During training, we first cluster discovery tasks by shared execution
needs identified in base-harness execution records. An \emph{evolution agent}
iteratively selects a cluster and parent harness, creates a subclass,
and evaluates it across the cluster's tasks. This feedback guides
subsequent revisions and accumulates as design and routing lessons.
This offers two advantages. \emph{Efficiency:} evaluating candidates on
their target clusters avoids rerunning the entire discovery set.
\emph{Generalization:} specialized changes leave the base and other
learned harnesses intact, so new tasks can use an appropriate harness
without inheriting unrelated changes.
At test time, \emph{polymorphism} allows the executor to run a selected
harness through a common execution interface. A \emph{harness router}
uses the task, an observable base trajectory, and routing lessons to
select a specialized harness or retain the base result. The hierarchy and lessons are frozen before deployment;
executor weights remain fixed in both phases.

Concurrent work also explores task-specific harness selection. Adaptive
Auto-Harness routes tasks among specialized Git
branches~\citep{liu2026adaptive}. In \ooh, inheritance shares code across
harnesses, and routing lessons learned during training guide selection
alongside class descriptions and observable base trajectories.

We evaluate \ooh on AutomationBench, Workspace-Bench, and Toolathlon-GYM,
where it achieves the highest primary score among the compared methods.

Our \textbf{contributions} are:
\begin{itemize}[nosep, before={\interlinepenalty=10000}]
  \item A harness representation combining shared code and specialized behavior
  through inheritance, method overrides, and new methods.
  \item An evolution framework that learns specialized harnesses and routing
  lessons from execution records to support task-specific selection.
  \item Strong empirical results across three long-horizon benchmarks,
  with extensive ablations validating individual components.
\end{itemize}

\section{Related Work}
\label{sec:related}
\paragraph{Self-evolving agents.}
Self-evolving agents use execution experience to improve prompts, memories,
skills, and other persistent components.
GEPA reflects on trajectories to optimize prompts \citep{agrawal2025gepa},
while ReasoningBank distills successful and failed experiences into reusable
memories \citep{ouyang2026reasoningbank}. A-Evolve considers deliberate
updates to persistent agent state \citep{lin2026aevolve}, and HASP represents
skills as executable functions that modify actions or supply corrective
context \citep{liu2026hasp}. These forms of adaptation provide building
blocks for harness evolution. \ooh evolves specialized harnesses
through inheritance and uses a harness router to select one for each task.

\paragraph{Automatic harness evolution.}
Recent work automates harness construction and revision through code search
and execution feedback, including Meta-Harness \citep{lee2026metaharness},
Agentic Harness Engineering \citep{lin2026ahe}, and HarnessEvolve
\citep{jiang2026harnessevolve}. These approaches primarily optimize a global
harness across tasks, although changes that benefit one task group may not
benefit another. HarnessDev benchmarks harness creation and evolution
\citep{wu2026harnessdev}, while EvoHarnessBench shows that expanding tools,
skills, and specialist agents can degrade previously effective behavior
\citep{ke2026evoharnessbench}. Concurrent work explores adaptation across
tasks and deployments: Adaptive Auto-Harness shares and specializes workspace
contents through Git branching and rebasing \citep{liu2026adaptive}, while EvolveNet
integrates adaptations from distributed deployments into a global harness
\citep{nie2026evolvenet}. At the representation level, NOOA provides a
programming model that expresses agents as native Python objects
\citep{furgale2026nooa}. \ooh evolves a hierarchy of executable harness
classes through inheritance, method overrides, and new methods. Its harness
router selects a harness using routing lessons accumulated during training,
class descriptions, the task, and an observable base trajectory.

\section{Method}
\label{sec:method}

\subsection{Problem setting and overview}
\label{sec:method:overview}
\begin{figure}[!ht]
  \centering
  \includegraphics[width=\linewidth]{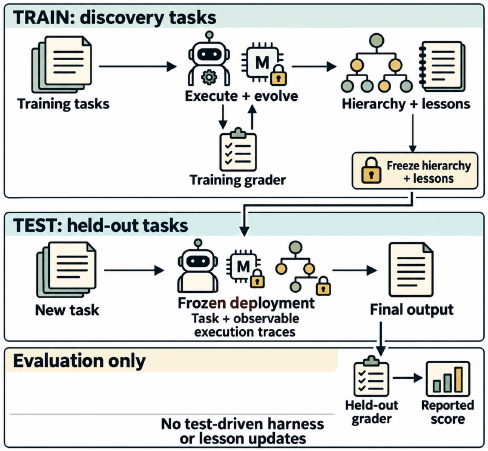}
  \caption{\textbf{Problem setting.}
  We use a train/test split: harnesses evolve on discovery tasks and are
  evaluated on disjoint held-out tasks.}
  \label{fig:setting}
\end{figure}

A harness implements the prompts, tools, and execution rules around an
\emph{executor}, the language model $M$ that carries out tasks
\citep{ning2026codeharness}. \ooh evolves this code on discovery tasks while
keeping $M$ fixed, then evaluates the frozen harnesses on disjoint held-out
tasks (Figure~\ref{fig:setting}). This differs from Meta-Harness's
TerminalBench-2 discovery experiment, which uses the same tasks for search
and final evaluation \citep{lee2026metaharness}. Separating discovery from
evaluation is essential for assessing generalization beyond the search set
\citep{wang2026rethinking}.

\begin{figure*}[!t]
  \centering
  \includegraphics[width=\textwidth]{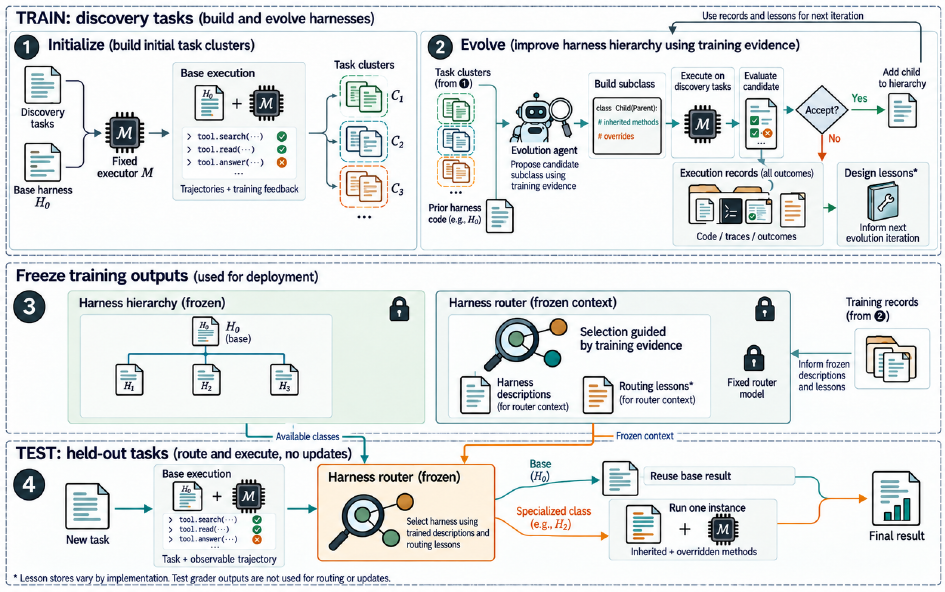}
  \caption{\textbf{OO-Harness training and test-time execution.}
  \textbf{(1)} Base execution records initialize task clusters.
  \textbf{(2)} The evolution agent builds and evaluates subclasses, updating
  lessons from execution feedback.
  \textbf{(3)} The resulting harness hierarchy, harness descriptions, and
  routing lessons are frozen for testing.
  \textbf{(4)} The harness router uses the task, observable base trajectory,
  and frozen routing lessons to select a specialized harness or reuse the
  base result.}
  \label{fig:method}
\end{figure*}

A \emph{\textbf{harness hierarchy}} $\mathcal{T}_H$ stores complete harness classes,
starting from the base $H_0$. Each child inherits its parent's methods and
can override existing methods or add new ones. \emph{Task clusters} $\mathcal{C}$ group
discovery tasks with related execution needs. For cluster $c$, let
$\mathcal{D}_c$ denote its discovery tasks and $H_c$ the harness currently
chosen for that cluster. Clusters organize task evidence; the hierarchy
organizes reusable harness code.
The \emph{\textbf{evolution agent}} $\pi$ constructs candidate subclasses from
discovery-task evidence, while the \emph{harness router} $\rho$ selects the
base or a specialized harness (Figure~\ref{fig:method}).
\ref{app:models-prompts},\ref{app:ooh-implementation} describe model-prompts/implementation details;
Table~\ref{tab:routing} compares routing inputs.

Execution records $\mathcal{E}$ store candidate code, discovery-task
trajectories, outcomes, and reflections from successful and unsuccessful
proposals. Where lesson stores are enabled, we summarize these records into
\emph{design lessons} $\mathcal{L}_{\pi}$ for code revision and
\emph{routing lessons} $\mathcal{L}_{\rho}$ for harness selection.
Design lessons describe useful changes and failed approaches; routing
lessons describe when a harness applies. Both accumulate during discovery
and remain frozen at deployment. The router's model weights stay fixed;
training changes the evidence that guides its decisions.

\subsection{Representing the harness hierarchy}
\label{sec:method:oop}
\paragraph{Interfaces and inheritance.}
Harness evolution operates on executable class definitions and supporting
source files. The agent can change prompts, tools, context management,
and execution control, and introduce additional methods. Where supported
by the scaffold, it can also configure middleware, file backends, and
subagents. The resulting class must remain compatible with the scaffold's
execution interface.

A subclass reuses inherited methods and overrides or adds methods for a
targeted change. Overrides can call parent methods through
\lstinline|super()|. Figure~\ref{fig:motivation} illustrates the hierarchy;
Listing~\ref{lst:node} shows inheritance in a saved harness lineage.

\paragraph{Execution within one instance.}
The harness router selects a complete class whose methods execute within
one instance; ancestors are not separate agent runs. Methods can condition
their behavior on runtime evidence, such as repeated tool errors or
unfinished output artifacts.

\subsection{Training initialization}
\label{sec:method:discovery}
We first run $H_0$ on discovery tasks $\mathcal{D}_{\mathrm{tr}}$ and retain
trajectories and evaluation feedback. We group tasks whose base executions
suggest that they may benefit from a shared harness change, yielding
clusters $\mathcal{C}$ and task sets $\mathcal{D}_c$. Each cluster records its
shared execution needs and starts with $H_c=H_0$; the initial hierarchy is
$\mathcal{T}_H=\{H_0\}$.

\subsection{Training iteration}
\label{sec:method:evolution}
\label{sec:method:accept}

At each iteration, the evolution agent works in a code workspace with
filesystem access to harness source, task clusters, and execution records.
It can inspect files, delegate analysis to subagents, and
write or edit candidate code. It selects a target cluster and parent class,
states the intended change, and implements an executable subclass.
Successful and unsuccessful proposals remain available for later inspection.

We execute the candidate $H'$ on $\mathcal{D}_c$ with the fixed model $M$.
Let $S_c(H')$ denote its cluster score and $S_c^{\mathrm{current}}$ the
stored score under current harness assignments.
The candidate's source parent need not be $H_c$.
Evaluation runs directly on the target cluster without routing,
\textbf{avoiding rerunning the full discovery set for every candidate}.

We accept $H'$ when $S_c(H') > S_c^{\mathrm{current}}$ and
benchmark-specific acceptance checks pass; otherwise, we reject it.
These checks consider task-level outcomes and execution feedback,
with a judging agent for ambiguous cases where configured.
Accepted classes enter $\mathcal{T}_H$ and can become the cluster's
selected harness $H_c$. All outcomes guide later revisions and lesson
updates. Improvement is required on the target cluster, not all tasks.
A parent remains reusable through inheritance and, if active, can still
be selected separately from its specialized children.

Repeated rejections lower a cluster's search priority, allowing evolution
to explore other task groups. Cluster assignments are also re-examined at
each iteration: execution feedback can motivate reassigning tasks or
splitting or merging clusters, including when repeated harness revisions
fail to improve a group.
Algorithm~\ref{alg:evolve} summarizes evolution;
Appendix~\ref{app:ooh-implementation} gives implementation details.

\begin{algorithm}[!t]
\caption{Harness evolution on discovery tasks 
}
\label{alg:evolve}
\small
\begin{algorithmic}[1]
\State \textbf{Input:} $H_0$, fixed $M$, tasks $\mathcal{D}_{\mathrm{tr}}$, budget $T$
\State Run $H_0$ on $\mathcal{D}_{\mathrm{tr}}$; store execution records
\State Cluster tasks from these records into $\{\mathcal{D}_c\}$
\State Initialize $\mathcal{T}_H$ and $H_c\gets H_0$
\For{$i=1,\ldots,T$}
  \State Review clusters; adjust task assignments if needed
  \State $\pi$ chooses a cluster $c$ and parent in $\mathcal{T}_H$
  \State Build subclass $H'$; override or add methods
  \State Run $H'$ on $\mathcal{D}_c$; collect scores and feedback
  \State Compare task outcomes; apply acceptance checks
  \State Retain accepted $H'$; update $H_c$ if preferred
  \State Update cluster priority from acceptance history
  \State Store outcomes; distill enabled lessons
\EndFor
\State \textbf{Output:} freeze $\mathcal{T}_H$, $\{H_c\}$, records, lessons
\end{algorithmic}
\end{algorithm}

\subsection{Test-time selection and execution}
\label{sec:method:deploy}
For a new task $t$, the harness router observes the base trajectory
$\omega(t)$ and selects
\[
H^\star=\rho\bigl(t,\omega(t),\{H_0\}\cup\{H_c\}_{c\in\mathcal{C}},\mathcal{L}_{\rho}\bigr).
\]
It uses class descriptions, applicability conditions, and frozen routing
lessons. If $H^\star=H_0$, we reuse the base result. Otherwise, we execute
one instance of $H^\star$ with its inherited and specialized behavior.
The hierarchy and stored evidence remain fixed across all held-out tasks.
Grader feedback supports evolution on training tasks only. At test time,
grader scores, verdicts, and hidden reference answers are unavailable to the
router or executor.

\section{Experiments}
\label{sec:exp}

\subsection{Evaluation setup}
\label{sec:exp:setup}
\paragraph{Benchmarks and metrics.}
We evaluate on AutomationBench~\citep{shepard2026automationbench} for
business workflows (300 discovery and 300 held-out tasks),
Workspace-Bench~\citep{tang2026workspacebench} for work across files and
applications (288 discovery and 100 held-out tasks, Lite-100), and
Toolathlon-GYM~\citep{zhang2026toolathlongym} for multi-tool task completion
(251 discovery and 252 held-out tasks). AutomationBench and Toolathlon-GYM
use deterministic graders to assess task completion; Workspace-Bench
evaluates outputs against task-specific rubrics.

We report Mean and strict task success (PASS) for all three benchmarks.
Mean denotes mean partial credit on AutomationBench, micro-averaged rubric
pass rate on Workspace-Bench, and mean reward on Toolathlon-GYM.
PASS requires reward at least $0.999$ on AutomationBench and Toolathlon-GYM,
or all rubrics satisfied on Workspace-Bench. The primary metric is Mean
for AutomationBench and Workspace-Bench, and PASS for Toolathlon-GYM.
All scores are reported as percentages.

\paragraph{Comparators.}
\label{sec:method:vsmeta}
We compare with the unevolved \textsc{base} harness; global harness
optimization using \mh~\citep{lee2026metaharness} and Agentic Harness
Engineering (\ahe; \citealp{lin2026ahe}); and specialized harnesses using
Adaptive Auto-Harness (\aah; \citealp{liu2026adaptive}). Our \ahe\ adaptation
iteratively edits a global harness from a framework-only bundle.
Since \ooh\ uses a base trajectory for routing, we also compare with
test-time scaling of the base harness to assess gains from additional
inference. Inspired by \citet{wang2026rethinking}, we evaluate parallel
best-of-$K$ selection ($K\in\{2,5\}$) and sequential self-refinement
($K=2$; \citealp{madaan2023selfrefine}). Both methods use observable
trajectories without grader feedback.

\paragraph{Evaluation and reporting.}
All methods use the same \texttt{gemini-3.5-flash} executor and follow each
benchmark's native execution protocol. We fix the evolution agent to Claude
Opus 4.8 across harness optimization methods for up to 40 iterations and use the same model for
test-time trajectory selection or guidance where required. The \ooh\ harness
router uses Claude Sonnet 5 (Appendix~\ref{app:prompt-interfaces}). Harnesses and
lessons remain frozen during held-out evaluation, and grader outputs are
used only for final scoring (\S\ref{sec:method:deploy}). For the main
comparison, we perform three independent executor runs per method and
benchmark and report mean $\pm$ standard deviation. \emph{Steps} is the mean number of steps in the finally selected
trajectory, excluding other rollouts and auxiliary selection calls.
\subsection{Main results}
\label{sec:exp:main}
\begin{table*}[t]
\caption{Held-out results with a frozen executor.
\emph{Mean}: mean partial credit on AutomationBench, micro-averaged rubric pass rate
on Workspace-Bench, and mean reward on Toolathlon-GYM.
\emph{PASS}: strict task success (reward $\geq 0.999$ on AutomationBench and Toolathlon-GYM;
all rubrics satisfied on Workspace-Bench).
The primary metric is Mean for AutomationBench and Workspace-Bench, and PASS for Toolathlon-GYM.
\emph{Steps} counts only the finally selected trajectory, excluding other rollouts and auxiliary selection calls.
Values are mean $\pm$ standard deviation over three runs; scores are percentages.
Bold marks each column's maximum score ($\uparrow$) or minimum steps ($\downarrow$).
$^{\dagger}$: supervised harness evolution with discovery-task feedback.
$^{\ddagger}$: unsupervised test-time scaling of the base harness.
Held-out grader feedback is unavailable to all methods.
}
\label{tab:main}
\centering
\small
\setlength{\tabcolsep}{3.5pt}
\renewcommand{\arraystretch}{1.12}
\resizebox{\textwidth}{!}{%
\begin{tabular}{lccccccccc}
\toprule
 & \multicolumn{3}{c}{AutomationBench} & \multicolumn{3}{c}{Workspace-Bench} & \multicolumn{3}{c}{Toolathlon-GYM} \\
\cmidrule(lr){2-4}\cmidrule(lr){5-7}\cmidrule(lr){8-10}
Method & Mean $\uparrow$ & PASS $\uparrow$ & Steps $\downarrow$ & Mean $\uparrow$ & PASS $\uparrow$ & Steps $\downarrow$ & Mean $\uparrow$ & PASS $\uparrow$ & Steps $\downarrow$ \\
\midrule
\textsc{base} & $61.62_{\pm0.15}$ & $25.44_{\pm0.51}$ & $31.27_{\pm0.13}$ & $25.45_{\pm2.98}$ & $3.67_{\pm0.58}$ & $34.41_{\pm1.34}$ & $71.87_{\pm0.65}$ & $31.15_{\pm0.28}$ & $40.50_{\pm0.99}$ \\
\hdashline
\rowcolor{cathdr}\multicolumn{10}{l}{\textit{Test-time scaling}}\\
Parallel ($K{=}2$)$^{\ddagger}$ & $61.46_{\pm2.38}$ & $24.67_{\pm2.83}$ & $31.64_{\pm0.23}$ & $32.54_{\pm4.30}$ & $5.67_{\pm1.15}$ & $41.62_{\pm2.56}$ & $74.85_{\pm1.45}$ & $31.55_{\pm1.96}$ & $39.94_{\pm0.32}$ \\
Parallel ($K{=}5$)$^{\ddagger}$ & $60.24_{\pm1.07}$ & $24.00_{\pm0.47}$ & $31.29_{\pm0.05}$ & $37.71_{\pm0.45}$ & $7.67_{\pm0.58}$ & $48.63_{\pm1.92}$ & $75.69_{\pm0.79}$ & $32.14_{\pm0.56}$ & $38.90_{\pm0.00}$ \\
Sequential ($K{=}2$)$^{\ddagger}$ & $59.29_{\pm0.25}$ & $25.00_{\pm1.41}$ & $31.86_{\pm0.12}$ & $27.93_{\pm2.13}$ & $3.33_{\pm2.52}$ & $32.91_{\pm2.07}$ & $73.53_{\pm1.67}$ & $33.33_{\pm2.81}$ & $40.97_{\pm1.32}$ \\
\hdashline
\rowcolor{cathdr}\multicolumn{10}{l}{\textit{Global harness optimization}}\\
\mh{}$^{\dagger}$ & $63.92_{\pm0.89}$ & $26.33_{\pm1.33}$ & $30.05_{\pm0.03}$ & $29.56_{\pm1.23}$ & $5.54_{\pm1.23}$ & $35.23_{\pm1.44}$ & $75.84_{\pm0.76}$ & $33.45_{\pm1.23}$ & $39.85_{\pm1.03}$ \\
\ahe{}$^{\dagger}$ & $64.18_{\pm0.91}$ & $28.11_{\pm1.84}$ & $30.61_{\pm0.18}$ & $38.43_{\pm4.32}$ & $7.78_{\pm0.87}$ & $34.59_{\pm1.21}$ & $74.48_{\pm1.23}$ & $36.23_{\pm1.19}$ & $41.02_{\pm0.89}$ \\
\hdashline
\rowcolor{cathdr}\multicolumn{10}{l}{\textit{Specialized harness families}}\\
\aah{}$^{\dagger}$ & $75.40_{\pm1.26}$ & $34.31_{\pm1.19}$ & $28.89_{\pm0.03}$ & $38.62_{\pm2.98}$ & $8.00_{\pm0.82}$ & $33.58_{\pm2.48}$ & $74.99_{\pm1.63}$ & $34.32_{\pm1.47}$ & $41.23_{\pm0.88}$ \\
\ooh{}$^{\dagger}$ & $\mathbf{77.41}_{\pm1.37}$ & $\mathbf{37.22}_{\pm2.14}$ & $\mathbf{27.99}_{\pm0.02}$ & $\mathbf{39.48}_{\pm3.21}$ & $\mathbf{8.33}_{\pm2.52}$ & $\mathbf{32.78}_{\pm1.97}$ & $\mathbf{77.18}_{\pm1.15}$ & $\mathbf{37.45}_{\pm0.98}$ & $\mathbf{37.25}_{\pm0.38}$ \\
\bottomrule
\end{tabular}%
}
\vspace{1pt}
\end{table*}

\paragraph{Strong empirical results.}
\ooh\ achieves the highest primary score on all three benchmarks
(Table~\ref{tab:main}), outperforming prior global harness optimization
approaches and \aah's specialized harnesses. These results support the
expressiveness of object-oriented programming as a natural fit for
code-based harness evolution: inheritance shares useful behavior, while
method overrides and additions specialize it for different tasks.
On AutomationBench, \ooh\ also has the lowest mean trajectory step count
among all compared methods.
Table~\ref{tab:main} also shows that test-time scaling with parallel
best-of-$K$ selection ($K\in\{2,5\}$) and sequential self-refinement
($K=2$) does not consistently improve performance. Additional inference
alone therefore does not guarantee gains, highlighting the value of
learning task-appropriate harnesses through our training pipeline.

\begin{figure}[!ht]
  \centering
  \includegraphics[width=\linewidth]{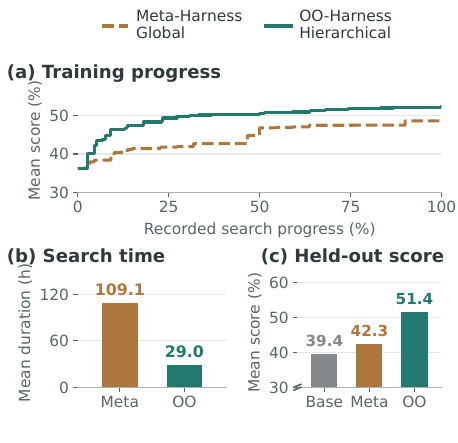}
  \caption{\textbf{Global versus hierarchical harness optimization.}
  Compared with \mh, \ooh\ shows (a) larger early training gains,
  (b) shorter recorded search time, and (c) higher held-out scores.
  Curves align runs by their fraction of recorded updates.}
  \label{fig:search-summary}
\end{figure}

\paragraph{Global versus hierarchical approaches.}
\label{sec:exp:adaptive}
We compare more closely with \mh, a prominent prior approach to global
harness optimization. During training, \mh\ evolves one global harness,
whereas \ooh\ builds a hierarchy of specialized harnesses through
cluster-specific evolution. Both freeze code before testing; \mh\ applies
one harness to every task, while \ooh\ selects from its learned hierarchy.
Figure~\ref{fig:search-summary} highlights three advantages of \ooh:
\textbf{training progress}, with larger early gains in the averaged
discovery curve (a); \textbf{training efficiency}, with shorter recorded
search duration (b); and \textbf{generalization performance}, with higher
held-out scores (c). Evaluating revisions on target clusters instead of
rerunning the full discovery set can reduce API calls and associated cost
per candidate. These results motivate hierarchical, adaptive harness
optimization for heterogeneous tasks.

\section{Analysis}
\label{sec:analysis}
In this section, we aim to understand the effectiveness of \ooh\ through
ablations of (1) task- and execution-informed harness routing
(\S\ref{sec:analysis:routing}) and (2) hierarchical harness optimization
through inheritance (\S\ref{sec:analysis:inheritance}). These comparisons
assess the roles of execution evidence and hierarchical code reuse.
We then conduct detailed case studies
of base-harness reuse, learned task clusters, and concrete harness changes
to relate execution needs to code and routing decisions
(\S\ref{sec:analysis:artifacts}). Finally, we construct an out-of-distribution (OOD) setting with
disjoint training and test application families, and examine Base reuse
on the selected test tasks (\S\ref{sec:analysis:ood}).

\subsection{Ablation studies}
\label{sec:analysis:ablations}
\subsubsection{Harness routing}
\label{sec:analysis:routing}
We compare two routers with the same frozen harness catalogue, executor,
grader, and held-out split. The \emph{task-only} router uses task descriptions
and metadata, requiring self-reported applicability confidence of at least
$0.7$ to select a specialist. The trajectory-based router also observes the
base execution. Neither receives held-out grader feedback.
Appendix~\ref{app:task-only-prompts} details the task-only prompt setup.

\begin{table}[t]
\centering
\small
\setlength{\tabcolsep}{3.5pt}
\resizebox{\linewidth}{!}{%
\begin{tabular}{@{}lrrr@{}}
\toprule
 & & \multicolumn{2}{c}{\shortstack{\ooh\\routing evidence}} \\
\cmidrule(lr){3-4}
Benchmark / metric & Base & \shortstack{Task\\only} &
\shortstack{Task +\\trajectory} \\
\midrule
Automation / Mean & $61.30$ & $64.88$ & $\mathbf{77.52}$ \\
Workspace / Mean & $24.57$ & ${40.03}$ & $\mathbf{40.56}$ \\
Toolathlon / PASS & $31.35$ & $33.33$ & $\mathbf{37.70}$ \\
\bottomrule
\end{tabular}%
}
\caption{\textbf{Routing ablation on a frozen \ooh\ catalogue.}
\emph{Task only} uses the task description and available task metadata;
\emph{Task + trajectory} also observes the base execution.
\emph{Base} executes without routing. Entries report primary scores (\%)
from single ablation runs, distinct from the repeated runs in
Table~\ref{tab:main}. 
}
\label{tab:routing}
\end{table}

Table~\ref{tab:routing} shows that the value of execution evidence depends
on the workload. Compared with task-only routing, trajectory-based routing
improves AutomationBench mean partial credit by $12.64$ points and
Toolathlon-GYM PASS by $4.37$ points. The gap on Workspace-Bench is only
$0.53$ points.
To understand why these gains differ, we further inspect the learned task
clusters and the harness changes developed for them. This inspection
suggests that the information needed for routing depends on how the
evolution agent groups tasks and specializes harnesses to address their
shared execution needs.
On Workspace-Bench, the learned cluster labels align with task types such as
document writing and spreadsheet editing
(Figure~\ref{fig:cluster-domains}b), making the
requested output a useful cue for harness selection. On AutomationBench and Toolathlon-GYM,
some harnesses address execution issues shared across domains or apps,
such as leaving required writes incomplete or failing to finalize within
the tool-call budget. The base trajectory helps reveal whether these
issues occurred, allowing the router to select a harness designed to
address them.
Appendix Figures~\ref{fig:before-after} and~\ref{fig:shared-ancestor}
illustrate document-writing and completion-check changes.

\subsubsection{Hierarchical evolution structure}
\label{sec:analysis:inheritance}
One central design choice of \ooh\ is evolution via inheritance. We
examine its contribution on Workspace-Bench with three ways of learning
harnesses. The routing procedure, executor, task splits, and nominal search
budget are the same across variants; only how harness code is learned
changes. \textbf{OO} extends earlier harnesses through subclasses.
\textbf{Flat} copies previously evolved code into a class derived directly
from the base, without inheriting from a specialized harness.
\textbf{Base-reset} starts from the base and adds new changes without
carrying forward previously evolved code. Each variant conducts a separate
search and is evaluated once on Lite-100.

\begin{figure}[!t]
  \centering
  \includegraphics[width=.94\linewidth]{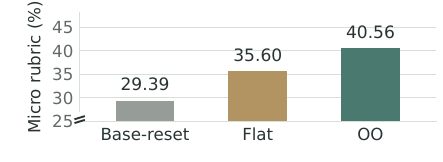}
  \caption{\textbf{Inheritance ablation on Workspace-Bench.}
  We find that learning through inheritance outperforms Flat and Base-reset.}
  \label{fig:inheritance-ablation}
\end{figure}

OO achieves the highest micro rubric pass of $40.56\%$, exceeding Flat by
$4.96$ points (Figure~\ref{fig:inheritance-ablation}). Flat also improves
over Base-reset, supporting the value of reusing useful code across
revisions. Since Flat also carries forward evolved code, its gap to OO
suggests a benefit from organizing revisions through inheritance.
Each variant evolves separately, so these results compare the harnesses
learned by each search strategy.

\subsection{Case studies}
\label{sec:analysis:artifacts}
\begin{figure*}[!t]
  \centering
  \includegraphics[width=\textwidth]{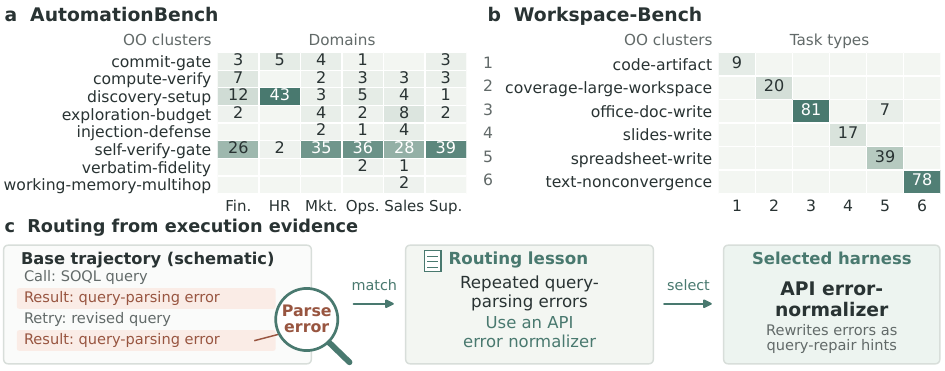}
  \caption{\textbf{Learned clusters and an example router lesson.}
  Panels (a,b) compare learned clusters (rows) with domains or task types
  (columns); cells count shared discovery tasks. AutomationBench clusters
  largely reflect execution needs, while Workspace-Bench clusters largely follow
  task types. Panel (c) schematically shows how repeated query-parsing
  errors in a Base trajectory match a saved routing lesson, leading to
  selection of an API error-normalizer harness. Blank cells are zero;
  Workspace-Bench column IDs match the numbered rows.}
  \label{fig:cluster-domains}
\end{figure*}

\paragraph{Reusing the base harness.}
\label{sec:analysis:base-routing}
In \ooh, a task is rerun with an evolved specialized harness only when
the router selects one; otherwise, the Base result is reused. In this case
study, we examine one AutomationBench run with 300 tasks, of which
134 ($44.7\%$) reuse Base and 166 select specialized harnesses.
Table~\ref{tab:base-reuse-audit}
compares the original Base performance of these two groups, before any
specialized execution. Tasks sent to specialists have lower Base mean
reward ($44.40\%$) than those selected for Base reuse ($77.15\%$);
strict PASS follows the same pattern. These results suggest that routing
directs specialization toward tasks with weaker Base executions, while
preserving stronger Base results through reuse. Routing lessons guide
this choice by linking observed execution problems to suitable harnesses
(Figure~\ref{fig:cluster-domains}c).

\begin{table}[t]
\centering
\small
\setlength{\tabcolsep}{4pt}
\resizebox{\linewidth}{!}{%
\begin{tabular}{@{}lrrr@{}}
\toprule
 & & \multicolumn{2}{c}{Original Base performance} \\
\cmidrule(lr){3-4}
Routing destination & Tasks & Mean reward & Strict PASS \\
\midrule
Reuse Base & 134 & 77.15 & 36.57 \\
Specialized harness & 166 & 44.40 & 14.25 \\
\bottomrule
\end{tabular}%
}
\caption{\textbf{Base performance by routing destination on AutomationBench.}
We find that tasks routed to specialized harnesses have substantially
weaker Base performance on average, while \ooh\ preserves stronger
Base executions through reuse.}
\label{tab:base-reuse-audit}
\end{table}

\paragraph{How task clusters reflect execution needs.}
The evolution agent can create and revise task clusters from discovery
executions, without a fixed domain or task-type taxonomy.
Figure~\ref{fig:cluster-domains} contrasts two learned organizations.
On AutomationBench, clusters capture failures and corrective needs visible
in trajectories: \texttt{commit-gate} targets missing writes, while
\texttt{compute-verify} groups tasks requiring arithmetic or time-based
derivations across five business domains. On Workspace-Bench, all six
learned cluster labels align with the evaluated \aah\ task-type groups,
including document writing and spreadsheet editing. Tasks requesting the
same output format often share backend requirements, so task-type clusters
can emerge naturally from execution evidence. The evolution agent can thus
group tasks by a shared failure or by a shared capability, depending on the
workload.

Routing lessons connect execution problems to harness capabilities.
In Figure~\ref{fig:cluster-domains}c, the router matches repeated query-parsing
errors to a saved lesson and selects an API error-normalizer that provides
query-repair instructions. Appendix~Table~\ref{tab:routing-lesson-example}
shows lessons that select a resource-manifest harness or preserve a clean
Base execution.

\paragraph{Concrete harness changes.}
Inheritance supports both shared behavior and targeted specialization.
On AutomationBench, the \texttt{discovery-setup} lineage extends a
missing-write completion check with a near-budget reminder and
tool-error guidance
(Appendix~Figure~\ref{fig:shared-ancestor}). Each node is a complete
executable harness. A parent supplies shared behavior and, if active,
can serve tasks that do not need a descendant's additional changes.
A document-backend specialization on Workspace-Bench
appears in Appendix~Figure~\ref{fig:before-after}.

\subsection{Base harness reuse in OOD settings}
\label{sec:analysis:ood}
\begin{figure}[!t]
  \centering
  \includegraphics[width=.94\linewidth]{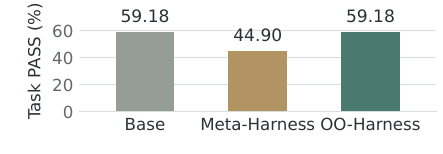}
  \caption{\textbf{Base reuse on the selected OOD test tasks.}
  After evolution on the disjoint training split, \ooh\ reuses Base on
  every test task, preserving its performance, while applying the learned
  \mh\ harness reduces PASS.}
  \label{fig:ood-transfer}
\end{figure}

To examine behavior in extreme OOD scenarios, we construct a setting with 254
Toolathlon-GYM training tasks and 49 Toolathlon
test tasks~\citep{li2025toolathlon}. Their named application families are
disjoint, making the test set largely out of distribution relative to the
training set (Appendix~\ref{app:ood-split}).
We evolve harnesses on this training split and compare \ooh\ against \mh.
At test time, \mh\ applies its learned global harness, whereas the \ooh\
router selects the general Base harness for every test task. As
Figure~\ref{fig:ood-transfer} shows, reusing Base preserves its performance,
while applying the \mh\ harness yields a lower PASS rate. Keeping Base
as a routing option allows \ooh\ to handle unfamiliar tasks without
requiring them to use a specialization learned on different application
families.

\section{Conclusion}
\label{sec:conclusion}
\ooh\ evolves specialized harnesses through inheritance and
routes tasks using execution evidence. Task clusters capture shared execution
needs, and inheritance allows specialized harnesses to share useful code. \ooh\ achieves the best performance among baselines. Ablations support inheritance and trajectory-informed
routing, while case studies illustrate the effectiveness of specialization and Base reuse.

\section*{Limitations}
\ooh\ relies on the APIs and execution environment provided by its host.
Service outages and rate limits can affect reliability, while interface
changes may require updates to evolved harnesses. Integrating private or
authenticated services also requires secure credential handling and
permission configuration. These dependencies motivate further work on
robust API integration and deployment across environments.

\clearpage
\begingroup
\small
\bibliographystyle{plainnat}
\bibliography{oo_harness}
\endgroup
\clearpage
\appendix
\section{Implementation and evaluation details}
\label{sec:appendix:implementation}
The implementations share the problem setting described in \S\ref{sec:method}, with
benchmark-specific settings described below.

\subsection{Model configuration}
\label{app:models-prompts}
For the main comparison across all three benchmarks, we use Anthropic's
Claude Opus 4.8 for evolution across harness-optimization methods~\citep{anthropic2026claudedocs}. We use Google's
\texttt{gemini-3.5-flash} as the fixed executor~\citep{google2026geminidocs} for all methods.
Execution follows each benchmark's native protocol, including
its agent scaffold, tool environment, execution limits, and evaluation
procedure. Harness evolution changes the code around the executor while
keeping its model weights fixed.

\subsection{Common training configuration}
\label{app:common-training}
We configure a budget of up to 40 evolution iterations for each
harness-optimization method on each benchmark, which we found sufficient
for harness search across the evaluated methods. Candidate evaluation and
selection use only discovery tasks. The resulting harnesses are frozen
before held-out evaluation.

\subsection{Benchmark implementations}
\label{app:benchmark-implementations}
We retain each benchmark's native task environment and evaluation procedure.
Task counts, discovery/held-out splits, and primary metrics are given in
\S\ref{sec:exp:setup}.

\paragraph{AutomationBench.}
Tasks execute business workflows in a simulated application environment.
The agent uses API search and request tools to inspect and modify an
initialized world state. We retain the native execution loop and tools.
Deterministic assertions evaluate the final state, yielding partial-credit
scores and strict task-completion verdicts.

\paragraph{Workspace-Bench.}
Tasks operate on files in a task workspace and require outputs such as
documents, spreadsheets, presentations, and code artifacts. We use the
DeepAgents scaffold with a local shell and filesystem backend. The
benchmark's agent-as-a-judge pipeline evaluates task-specific rubrics over
the resulting outputs. We report micro-averaged rubric pass and full-task
pass, which requires satisfying every rubric for a task.

\paragraph{Toolathlon-GYM.}
Tasks combine tools and application services exposed through configured
MCP servers. The native runner initializes each task's workspace and
service connections, then executes the agent through the benchmark's
containerized evaluation pipeline. Task-specific graders assess completion
and produce the evaluation records used to compute mean reward and strict
PASS.

\subsection{\ooh\ implementation}
\label{app:ooh-implementation}
This section describes the code workspace, prompts, and harness interface
used to implement \ooh\ (\S\ref{sec:method}).

\subsubsection{Evolution workspace and code authoring}
\label{app:evolution-workspace}
We run the evolution agent as a coding agent with filesystem access to
harness interfaces, existing source code, task clusters, discovery
trajectories, evaluation feedback, and prior reflections. It can search
and read these files, delegate evidence inspection to subagents, and write
executable candidate subclasses at designated paths. AutomationBench and
Workspace-Bench also support shared helper modules. The resulting code is
loaded, validated, and evaluated by the evolution procedure.

The Workspace-Bench authoring interface exposes file-reading, search,
writing, editing, and subagent tools. Its harness interface separately
supports tools, middleware, file backends, and subagent configurations.
For example, the \texttt{subagents()} method supplies subagents used during
benchmark task execution, while the evolution agent's subagents inspect
training evidence. The available modifications are determined by each
benchmark's harness interface.

\subsubsection{Evolution and routing prompts}
\label{app:prompt-interfaces}
\paragraph{Evolution prompt.}
The evolution agent receives the harness interface, current hierarchy,
target cluster or available clusters, and paths to discovery-task
trajectories, evaluation feedback, and prior reflections. A condensed
prompt is:
\par\noindent\begin{minipage}{\linewidth}
\begin{lstlisting}[basicstyle=\ttfamily\footnotesize]
Read the training trajectories and prior changes.
Use subagents to inspect the execution evidence.
Identify one shared, harness-fixable mechanism.
Write one subclass of the selected parent; call
super() to preserve the inherited behavior.
Do not hardcode task IDs, answers, or filenames.
Return a JSON manifest describing the change.
\end{lstlisting}
\end{minipage}
\par\smallskip
AutomationBench and Workspace-Bench bind the parent as
\texttt{\_Parent}; Toolathlon uses \texttt{\_Base}. AutomationBench
allows add, merge, and split proposals and lets the teacher choose a
cluster; Workspace-Bench supplies a target cluster and parent. Toolathlon
also provides its design principles and requires an observable applicability
\texttt{gate}. Supplied lesson text is capped at 6,000 characters in all
three implementations. The default proposal timeout is 2,400 seconds, separate from
the executor rollout budget. A representative manifest is:
\par\noindent\begin{minipage}{\linewidth}
\begin{lstlisting}[basicstyle=\ttfamily\footnotesize]
{"op": "add", "cluster": "<training cluster>",
 "parent": "<existing node or base>",
 "name": "<candidate name>",
 "seam": "<modified interface>",
 "hypothesis": "<shared failure and mechanism>"}
\end{lstlisting}
\end{minipage}
\par\smallskip
Toolathlon adds \texttt{gate}; Workspace-Bench adds \texttt{changes} and
names the seam as \texttt{system\_prompt}, \texttt{tools},
\texttt{middleware}, \texttt{subagents}, or \texttt{backend}.
Merge and split proposals identify the affected clusters or an observable
split criterion. The template files give the exact benchmark-specific
schemas; the evolution procedure validates candidates and determines acceptance from
training evaluations.

\paragraph{Routing prompt.}
The router receives the task, an observable Base-execution summary, and a
frozen harness catalogue, together with routing lessons where supplied.
We use Claude Sonnet 5 for routing in the main comparison. The condensed
AutomationBench prompt below illustrates the selection instruction:
\par\noindent\begin{minipage}{\linewidth}
\begin{lstlisting}[basicstyle=\ttfamily\footnotesize]
Choose one available harness or base using
only the Base query, observable trajectory,
and retrieved training context.
Choose base when no mechanism clearly fits
or a learned anti-pattern applies.
AVAILABLE HARNESSES: <frozen catalogue>
RETRIEVED CONTEXT: <retrieval response>
BASE QUERY + DIGEST: <observable evidence>
Return only JSON with choice, rule, and why.
\end{lstlisting}
\end{minipage}
\par\smallskip
Held-out grades and reference answers are excluded from the routing input.

\paragraph{Inputs supplied to the router.}
We summarize each benchmark's task metadata and observable Base execution
as a compact prompt record. These records provide cues for matching a task
to a learned cluster, such as the tools involved, encountered errors, and
required output format.

AutomationBench supplies the task query and a bounded digest of Base tool
calls and results, including write counts and execution-error patterns.
A first prompt retrieves relevant frozen routing rules; a second prompt
selects one available harness. The final response has the following schema:
\par\noindent\begin{minipage}{\linewidth}
\begin{lstlisting}[basicstyle=\ttfamily\footnotesize]
{"choice": "<harness-id|base>",
 "rule": "<rule-id or empty>",
 "why": "<one sentence>"}
\end{lstlisting}
\end{minipage}
\par\smallskip
For Toolathlon, each record contains the task identifier, tool families,
number of recorded history events, frequent tool or event types, and
execution-error categories. These fields summarize the tool environment,
interaction history, and errors observed during Base execution.

For Workspace-Bench, each record contains benchmark-provided task metadata:
persona and difficulty labels, requested output-file extensions, and the
number of links in the file-dependency graph. It also includes Base
assistant-turn and tool-call counts. Output types and file dependencies
describe the requested artifact and workspace, while the execution counts
summarize Base's activity. The router can therefore match both task
requirements and execution patterns to the learned cluster descriptions.

Both routers receive the frozen cluster menu alongside these records;
Toolathlon also receives frozen routing lessons. They return one assignment
per task:
\par\noindent\begin{minipage}{\linewidth}
\begin{lstlisting}[basicstyle=\ttfamily\footnotesize]
{"task_to_cluster": {
  "<task-id>": "<cluster-id|leave-on-base>"
}}
\end{lstlisting}
\end{minipage}
\par\smallskip
\paragraph{Router invocation.}
The router is called with agent tools disabled and receives only the
prepared task records and frozen routing context.

\paragraph{Router settings.}
The following code defaults specify limits on prompt length, batch sizes,
model choices, and call timeouts.
In the AutomationBench reference router, the query is limited to 600
characters and the digest processes at most 60 observable message entries;
the final digest is capped at 5,000 characters. Each compact routing lesson
contributes at most 180 characters of lesson text and each available-harness
description at most 120 characters. Retrieval requests a synthesis of at most ten lines.
Defaults are Sonnet 5 for retrieval and Opus 4.8 for selection, with
400- and 1,200-second timeouts respectively.
Toolathlon defaults to batches of 25 tasks, up to twelve frozen routing
rules capped at 4,000 characters in total, and a 1,800-second call timeout.
Workspace-Bench supplies the requested cohort's feature table in one call,
with a 1,200-second timeout. Its metadata fallback uses dependency counts and output extensions;
Toolathlon's invalid-menu fallback selects \texttt{leave-on-base}.
Both allow router-model overrides, otherwise using the proposer model
(default Opus 4.8). The case study in Table~\ref{tab:base-reuse-audit}
uses Opus 4.8 for both retrieval and selection.

\subsubsection{Inheritance example}
Listing~\ref{lst:node} illustrates how a specialized harness retains an
earlier capability while adding a new one. In this saved Toolathlon
lineage, parent s7 detects formula-like writes to \texttt{.xlsx} workbooks
and adds a reminder to save computed values rather than unevaluated
formula strings.
Child s30 inherits this guidance and adds a check of percentage-labelled
cells in saved workbooks before finishing. The check can request one
correction pass on the identified cells; Appendix~\ref{app:cs:ta} details
its trigger and unit convention.

The child does not redefine \seam{manage\_context}, so the parent's context
rule remains active. Its \seam{verify\_before\_finish} method calls
\lstinline|super()| to preserve inherited completion checks before applying
its own. The selected child combines these capabilities in one execution.
This illustrates incremental specialization: evolution can add a targeted
behavior while reusing an earlier mechanism without copying its code.

\begin{lstlisting}[float=*t,caption={Inherited context guidance with an added completion check (Toolathlon s7 $\rightarrow$ s30). Display names and helper calls are shortened; the method names match the implementation.},label={lst:node}]
class FormulaLiteralHarness(BaseHarness):
  def manage_context(self, messages, state):
    messages = super().manage_context(messages, state)
    return pin_literals_if_needed(messages, state)

class PercentCheckHarness(FormulaLiteralHarness):
  # Inherits manage_context from its parent.
  def verify_before_finish(self, messages, state):
    parent = super().verify_before_finish(messages, state)
    if parent:
      return parent
    return check_saved_percent_values(messages, state)
\end{lstlisting}

\subsection{Baseline method implementations}
\label{app:baseline-implementations}
All baselines use the task splits and fixed executor in
\S\ref{sec:exp:setup}. Harness optimization uses discovery-task feedback;
test-time trajectory selection and refinement use observable execution
records without grader feedback.

\paragraph{Base.}
Base uses the unevolved benchmark-specific harness, including its prompts,
tools, and execution loop. It performs no harness optimization or selection
among specialized harnesses.

\paragraph{Meta-Harness.}
\mh~\citep{lee2026metaharness}\footnote{\url{https://github.com/stanford-iris-lab/meta-harness}} searches for one global harness using
discovery-task evaluations. The evolution agent inspects earlier candidates
and execution feedback, proposes revised harness code, and evaluates the
resulting program. We freeze the selected global harness and apply it to
every held-out task.

\paragraph{AHE.}
\ahe~\citep{lin2026ahe}\footnote{\url{https://github.com/china-qijizhifeng/agentic-harness-engineering}} begins with framework-source
exploration and maintains an editable workspace of framework components.
A debugger analyzes discovery trajectories; the evolution agent uses these
reports and the change history to revise components, retain useful changes,
or roll back unsuccessful edits. The selected global snapshot is frozen
for held-out evaluation.

\paragraph{AAH.}
\aah~\citep{liu2026adaptive}\footnote{\url{https://github.com/A-EVO-Lab/a-evolve/tree/release/adaptive-auto-harness}} organizes tasks into groups
using task descriptions and metadata, and improves a separate harness for
each group. In our benchmark setup, AutomationBench groups correspond to business domains;
Workspace-Bench groups use output type and workspace requirements; and
Toolathlon-GYM groups follow application families. Each specialist is revised as a separate harness implementation. At deployment,
task descriptions or metadata determine which frozen specialist to use,
with Base used when no specialist is available.

\paragraph{Parallel sampling.}
For $K=2$ and $K=5$, we obtain independent Base executions of the same task.
A Claude Opus 4.8 selector compares their observable trajectories and
chooses one result. Neither reference answers nor grader feedback is
available to the selector.

\paragraph{Sequential refinement.}
For $K=2$, Claude Opus 4.8 summarizes process errors visible in the first
Base execution. These task-local notes guide a second Base execution,
without access to held-out grader feedback or reference answers.

\subsection{Ablation settings}
\label{app:ablation-settings}
These experiments isolate the roles of routing evidence and inheritance
in \ooh\ (\S\ref{sec:analysis:ablations}).

\subsubsection{Harness routing}
\label{app:task-only-prompts}
The routing ablation keeps the harness catalogue, executor, grader, and
held-out split fixed, varying only the routing input (Table~\ref{tab:routing}).
Task-only routing uses task descriptions and metadata; trajectory-based
routing also observes the Base execution. The task-only prompt setups
are given below.

\paragraph{Confidence gate.}
The router is asked for a specialist identifier and applicability confidence
$c\in[0,1]$. For a valid identifier and numeric confidence, the routing procedure
selects the specialist when $c\geq0.7$, otherwise Base. Confidence is
self-reported, not a calibrated success probability or a grader score.

\paragraph{AutomationBench.}
The prompt includes specialist identifiers with mechanisms truncated to
130 characters, and task identifiers, domains, and user triggers (up to
300 characters each), in batches of 25 tasks:
\begingroup
\begin{lstlisting}[basicstyle=\ttfamily\footnotesize]
Route each task to ONE specialist (or base)
using ONLY the task description/trigger.
You have no execution trace, tool output,
or score. Each specialist fixes a specific
failure mode of the base agent.

SPECIALISTS (id | mechanism): <catalogue>
TASKS (id | domain | trigger): <task batch>

Pick a specialist only when the description
clearly fits it; otherwise choose base.
Return a confidence in [0,1]. Output only JSON:
{"<task-id>": {"cluster": "<id-or-base>",
               "confidence": <0..1>}, ...}
\end{lstlisting}
\par\endgroup
Missing entries or unknown identifiers select Base; missing confidence
defaults to zero.

\paragraph{Workspace-Bench.}
Catalogue entries contain cluster identifiers, routing signatures, and
addressed failures. Each task supplies its identifier, persona, difficulty,
output-file extensions, dependency-edge count, tested capabilities (up to
80 characters), and task text (up to 500 characters). Batches default to
25 tasks:
\begingroup
\begin{lstlisting}[basicstyle=\ttfamily\footnotesize,breakatwhitespace=true,breakindent=1em]
Assign each new task to ONE cluster in the
fixed catalogue whose description best
matches it, using only the task description
and metadata. No execution trace, score,
or grader feedback is provided.

CLUSTERS (id | routing signature | failure):
<frozen catalogue>
TASKS (description + metadata): <task batch>

If no cluster matches well, or confidence
is below 0.7, choose leave-on-base.
Output only JSON: {"routes": {"<task-id>": {"cluster": "<cluster-id|leave-on-base>", "confidence": <0..1>}, ...}}
\end{lstlisting}
\par\endgroup
Missing or invalid identifiers invoke a task-metadata fallback based on
output-file type and dependency count; missing confidence defaults to one.

\paragraph{Toolathlon-GYM.}
The following reconstructed template uses task descriptions and declared
app/tool metadata with the frozen harness catalogue and the same confidence
gate. Catalogue entries describe each specialist's mechanism and
applicability conditions:
\begingroup
\begin{lstlisting}[basicstyle=\ttfamily\footnotesize]
Route each Toolathlon-GYM task to ONE
specialist in the fixed catalogue or
leave-on-base, using only its description
and declared app/tool metadata. No Base
trajectory, tool output, or grader feedback
is provided.

SPECIALISTS (id | mechanism | applicability):
<frozen catalogue>
TASKS (id | description | apps/tools):
<task batch>

Choose a specialist only when the task
requirements clearly match its capability.
Do not assume that an execution error has
occurred. If no specialist clearly fits or
confidence is below 0.7, choose leave-on-base.
Return each task exactly once. Output only JSON:
{"routes": {"<task-id>": {
  "cluster": "<cluster-id|leave-on-base>",
  "confidence": <0..1>}, ...}}
\end{lstlisting}
\par\endgroup

\subsubsection{Hierarchical structure}
\label{app:inheritance-ablation}
This ablation examines how an evolution step builds on previously learned
harness code. We compare three separate searches on Workspace-Bench, each
learning specialized harnesses for task clusters:
\begin{itemize}[leftmargin=*,itemsep=4pt,topsep=4pt,parsep=0pt]
\item \textbf{Base-reset (no code carry-over).}
Each candidate derives directly from Base. The evolution agent is asked
to implement one new mechanism without incorporating mechanisms from
earlier evolved nodes. Previous execution evidence and reflections remain
available to guide the proposal. Thus, each proposal starts afresh in its
code implementation while retaining the search's diagnostic history.

\item \textbf{Flat (copy-based reuse).}
Each candidate also derives directly from Base. When the target cluster
already has a deployed harness, the evolution agent is instructed to read
its source, copy all its accepted mechanisms into a new file, and then
add one new mechanism. The candidate cannot import, delegate to,
or inherit from other evolved nodes. Reuse therefore occurs through code
copied into the candidate, whose retained behavior is implemented within
the module.

\item \textbf{OO (inheritance).}
The target cluster's currently deployed harness becomes the candidate's
parent; Base is used when the cluster has no specialist. The loader binds
this parent class to \texttt{\_Parent}. Methods that the candidate does not
override remain inherited, while an override can call \texttt{super()} to
compose the parent's behavior with a new change. The evolution agent can
inspect prior source, execution evidence, and reflections, then add a
specialized behavior without needing to copy the parent's implementation.
\end{itemize}

\paragraph{Shared setup and evaluation.}
The variants use the same discovery split, initial task clusters, and
initial Base execution records. They share the procedures for choosing a target
cluster, scoring candidates, accepting revisions, and routing held-out
tasks, with the executor and grader fixed. Flat and Base-reset keep the
shared helper library read-only during their searches, so candidate
changes are written into the new node module.

Each search learns from the 288-task discovery split and freezes its
resulting catalogue for one held-out evaluation on Lite-100. The routing
procedure is shared, while the learned harnesses and resulting task
assignments can differ across searches. Figure~\ref{fig:inheritance-ablation}
reports micro rubric pass for the three resulting systems. Base-reset
versus Flat examines carrying forward evolved code; Flat versus OO
examines organizing that reuse through inheritance.

\subsection{OOD evaluation configuration}
\label{app:ood-split}
We use application metadata to construct the OOD split in
\S\ref{sec:analysis:ood}, retaining 254 Toolathlon-GYM tasks and 49 Toolathlon tasks. We note that the emphasis of Toolathon(-GYM) is on using applications and that different applications could require substantially different skills, tools, or workflows. The curated application distribution is as follows:
\begin{itemize}[leftmargin=*,nosep,topsep=3pt]
  \item \textbf{Training applications:} Google Workspace (Forms,
  Calendar, and Sheets), HowToCook, Notion, Snowflake, and WooCommerce.
  \item \textbf{Test applications:} Amazon, Canvas, Git/GitHub,
  Kubernetes, literature research (arXiv and Scholarly/Google Scholar),
  MySQL, 12306, Redis, Yahoo Finance, and YouTube.
\end{itemize}
As a result, these application families do not overlap and offer a good controlled experiment setting.

\subsection{Artifact sources and resource reporting}
\label{app:artifacts-resources}
\paragraph{Sources and terms.}
The AutomationBench repository distributes Zapier's original contributions
under the MIT license, with a separate scope notice for third-party API
structures.\footnote{\url{https://github.com/zapier/AutomationBench/blob/main/LICENSE}}
Workspace-Bench's code is MIT-licensed.\footnote{\url{https://github.com/OpenDataBox/Workspace-Bench/blob/main/LICENSE}}
Its full-dataset card lists Apache-2.0; the inspected Lite card does not
separately specify a license.\footnote{\url{https://huggingface.co/datasets/Workspace-Bench/Workspace-Bench};
\url{https://huggingface.co/datasets/Workspace-Bench/Workspace-Bench-Lite}}
Toolathlon-GYM's root code license is Apache-2.0.\footnote{\url{https://github.com/eigent-ai/toolathlon_gym/blob/main/LICENSE}}
The original Toolathlon repository and task-data card do not specify a
standard license.\footnote{\url{https://github.com/hkust-nlp/Toolathlon};
\url{https://huggingface.co/datasets/hkust-nlp/Toolathlon}}
Task files, bundled components, and external services may carry separate
terms that also govern redistribution.

\paragraph{API access and AI assistance.}
We access the language models through hosted APIs, using Google Cloud
Vertex AI as the backend. AI tools assisted manuscript revision and
polishing, code inspection, and figure preparation.

\paragraph{Resources.}
Across all benchmarks and methods, the main computational expense is
hosted-model API usage for task execution and, where applicable, harness
evolution, routing, and model-based evaluation.

\section{Case studies of specialized harnesses}
\label{app:casestudy}
\raggedbottom
These cases show how \ooh\ turns execution needs into concrete code
changes and routing choices. We examine document generation on
Workspace-Bench, shared completion checks and routing lessons on
AutomationBench, and targeted runtime checks on Toolathlon-GYM.

\subsection{Workspace-Bench: generating Word documents}
\label{app:cs:wb}
\begin{figure}[!t]
  \centering
  \includegraphics[width=\linewidth]{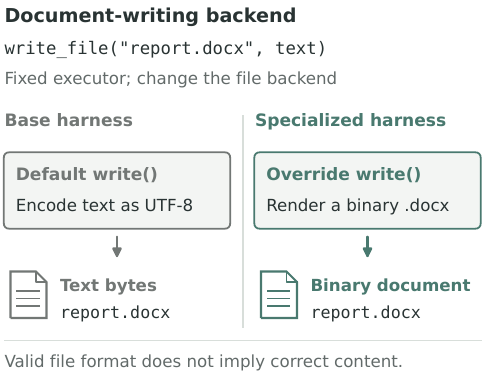}
  \caption{\textbf{Generating a Word document through the file backend.}
  For the same illustrative \texttt{write\_file} call, the base backend
  saves text under a \texttt{.docx} filename; the specialized backend
  renders the text into a binary document.}
  \label{fig:before-after}
\end{figure}

Document-writing tasks require both suitable content and a valid output
file. Writing plain text to a \texttt{.docx} path does not produce a Word
document. The \texttt{office-doc-write} harness addresses this problem
by wrapping the file backend and rendering supported text into a binary
Word document (Figure~\ref{fig:before-after}).

The specialized backend handles direct document writes and can also
create a missing requested document from a text draft. In the latter
case, it preserves the original text write and leaves any existing
target document untouched. If rendering fails, execution falls back to
the parent backend. This capability handles the file format; the
benchmark separately evaluates whether the content satisfies the task.

The requested output format is often visible in the task description,
giving the router a useful cue for selecting this capability. This is
consistent with the small difference between task-only and
trajectory-based routing on Workspace-Bench.

\subsection{AutomationBench: inherited checks and routing lessons}
\label{app:cs:ab}
\paragraph{Adding execution checks through inheritance.}
Business workflows often require changes to application state, but an
execution may finish after searching or reading without making an
update. An ancestor (s9) addresses this problem with a completion
check: when the task requires a write but no mutating API call has
occurred, it can request another turn.

\begin{figure}[!htbp]
  \centering
  \includegraphics[width=\linewidth]{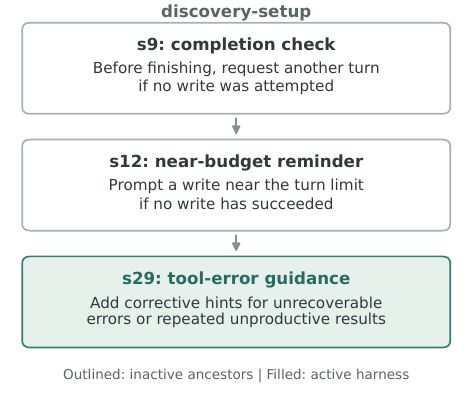}
  \caption{\textbf{Adding execution checks through inheritance.}
  Successive nodes add a missing-write completion check, a near-budget
  reminder, and tool-error guidance. Arrows show inheritance.
  The filled node s29 is active; s9 and s12 are inactive ancestors
  that supply inherited behavior.}
  \label{fig:shared-ancestor}
\end{figure}

Figure~\ref{fig:shared-ancestor} shows successive extensions within the
\texttt{discovery-setup} lineage. Building on s9's missing-write
completion check, s12 adds a reminder near the turn limit when no
write has succeeded. Its child s29 retains both mechanisms and adds
guidance for unrecoverable errors and repeated unproductive tool results.

\begin{table*}[t]
\centering
\small
\setlength{\tabcolsep}{5pt}
\begin{tabular}{@{}>{\raggedright\arraybackslash}p{0.34\textwidth}
                  >{\raggedright\arraybackslash}p{0.37\textwidth}
                  >{\raggedright\arraybackslash}p{0.21\textwidth}@{}}
\toprule
Base execution & Routing lesson & Router choice \\
\midrule
A lead-scoring execution makes no writes. HubSpot and Intercom return
401 errors indicating no connected account; a Lead lookup returns 404.
The final response is empty.
& Repeated connection errors and failed resource lookups suggest that
information about connected services and available resources may help.
& Resource-manifest harness. \\
\addlinespace
Another execution completes three writes, has no errors, and finishes
cleanly.
& Successful writes and a clean finish support keeping the Base result.
& Reuse Base. \\
\bottomrule
\end{tabular}
\caption{\textbf{Execution evidence behind two routing choices.}
Examples from an earlier AutomationBench workflow pair observed Base
behavior with the applicable routing lesson and the recorded choice.}
\label{tab:routing-lesson-example}
\end{table*}

\begin{figure*}[!t]
  \centering
  \includegraphics[width=.80\textwidth]{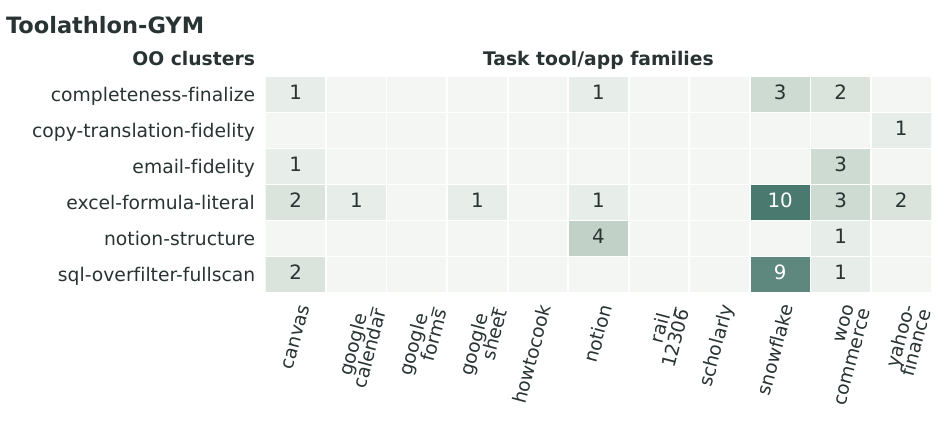}
  \caption{\textbf{Learned task clusters on Toolathlon-GYM.}
  Rows are learned clusters and columns are evaluated \aah\ app groups.
  Cells count shared discovery tasks, showing how a cluster can span
  several applications. Blank cells are zero; Base and unclustered tasks
  are omitted.}
  \label{fig:cluster-heatmap}
\end{figure*}

\paragraph{Using execution evidence to choose a harness.}
\label{app:error-trigger}
\label{app:routing-lesson-example}
Routing lessons connect observed execution problems to harness
capabilities. For example, a lesson recommends an API error-normalizer
when an execution repeatedly encounters query-parsing errors. This
harness turns error feedback into instructions for repairing the query
(Figure~\ref{fig:cluster-domains}c).

Table~\ref{tab:routing-lesson-example} gives two recorded routing choices.
An execution with no writes and repeated service errors is routed to a
resource-manifest harness, which supplies connected-service and resource
information to guide API exploration. A clean execution with successful
writes reuses Base. These examples show how lessons can select a
relevant intervention or preserve a clean Base execution.

\subsection{Toolathlon-GYM: refining when and what to check}
\label{app:cs:ta}
The learned clusters can span multiple applications
(Figure~\ref{fig:cluster-heatmap}). For example,
\texttt{completeness-\allowbreak finalize} includes tasks using Canvas, Notion,
Snowflake, and WooCommerce. The following examples show how evolution
refines a shared behavior and adds a check for a particular output.

\paragraph{Refining when to compress tool output.}
An ancestor of the \texttt{completeness-\allowbreak finalize} harness shortens
unusually long tool responses to limit the amount of text in the
execution history. Its child retains this behavior but adds an early-finish
condition: after an early completion attempt, any subsequent tool
outputs bypass compression. This allows the child to narrow when an
inherited intervention applies, using progress observed during execution.

\paragraph{Checking the saved workbook.}
The \texttt{excel-formula-literal} harness adds a file check to inherited
guidance about writing computed values to spreadsheets
(Listing~\ref{lst:node}). Before finishing, it reads the saved workbook
and looks for cells with percentage labels and values strictly between
zero and one. It can then request one correction pass, listing the
identified cells and proposed values multiplied by 100.

This rule assumes that the task expects percentage points: for example,
20 rather than 0.2 for twenty percent. The instruction asks the executor
to change only the listed cells and preserve the rest of the workbook.
The harness therefore uses the contents of the produced file to decide
whether to request a targeted correction before finishing.

\end{document}